\documentclass{article}

     \usepackage[preprint]{neurips_2021}

\usepackage[utf8]{inputenc} 
\usepackage[T1]{fontenc}    
\usepackage{hyperref}       
\usepackage{url}            
\usepackage{booktabs}       
\usepackage{amsfonts}       
\usepackage{nicefrac}       
\usepackage{microtype}      
\usepackage{xcolor}         

\usepackage{graphicx}
\usepackage{caption}
\usepackage{subcaption}
\usepackage{hyperref}
\usepackage{xcolor}
\usepackage{url}
\usepackage{arydshln} 
\usepackage{amssymb} 
\usepackage{booktabs}
\usepackage{float}
\usepackage{footnote}
\usepackage{multirow}
\usepackage{tikz}
\usepackage{amsmath}
\usepackage{color}
\usepackage{pgfplots}
\usepackage{pgf-umlsd}
\usepackage{ifthen}
\usepackage{forest}
\usepackage[misc]{ifsym}
\usepackage{algorithm}

\usepackage{algpseudocode}
\usepackage{amssymb}
\usepackage{wrapfig}
\usepackage{flushend}
\usepackage{subcaption}

\algtext*{EndWhile}
\algtext*{EndIf}
\algtext*{EndFor}

\title{Guided Evolution for Neural Architecture Search}

\author{%
  Vasco Lopes\\
  NOVA Lincs, Universidade da Beira Interior\\
  \texttt{vasco.lopes@ubi.pt} \\
   \And
   Miguel Santos \\
   NOVA Lincs, Universidade da Beira Interior \\
   \texttt{miguel.santos@ubi.pt} \\
   \And
   Bruno Degardin \\
   Universidade da Beira Interior \\
  \texttt{bruno.degardin@ubi.pt} \\
  \And
   Luís~A.~Alexandre \\
   NOVA~LINCS,~Universidade~da Beira~Interior\\
  \texttt{luis.alexandre@ubi.pt} \\
}
\newcommand{\matr}[1]{\mathbf{#1}} 
\begin{document}

\maketitle

\begin{abstract}
Neural Architecture Search (NAS) methods have been successfully applied to image tasks with excellent results. However, NAS methods are often complex and tend to converge to local minima as soon as generated architectures seem to yield good results. In this paper, we propose G-EA, a novel approach for guided evolutionary NAS. The rationale behind G-EA, is to explore the search space by generating and evaluating several architectures in each generation at initialization stage using a zero-proxy estimator, where only the highest-scoring network is trained and kept for the next generation. This evaluation at initialization stage allows continuous extraction of knowledge from the search space without increasing computation, thus allowing the search to be efficiently guided. Moreover, G-EA forces exploitation of the most performant networks by descendant generation while at the same time forcing exploration by parent mutation and by favouring younger architectures to the detriment of older ones. Experimental results demonstrate the effectiveness of the proposed method, showing that G-EA achieves state-of-the-art results in NAS-Bench-201 search space in CIFAR-10, CIFAR-100 and ImageNet16-120, with mean accuracies of 93.98\%, 72.12\% and 45.94\% respectively.

\end{abstract}

\section{Introduction}
Convolutional Neural Networks (CNNs) have been extensively applied with success to a panoply of tasks, from image classification \citep{deng2014deep,goodfellow2016deep}, to semantic segmentation \citep{garcia2018survey}, text analysis \citep{DBLP:conf/eacl/SchwenkBCL17}, amongst many others \citep{khan2020survey}. Their inherent capability of feature extraction allows CNNs to be easily applied and transferred to different problems. Over the years, several brilliantly and carefully designed architectures have incrementally out-performed the state-of-the-art by proposing novel components and mechanisms, such as skip and residual connections, faster and less size intensive operations and attention mechanisms \citep{DBLP:conf/nips/KrizhevskySH12,DBLP:conf/cvpr/SzegedyLJSRAEVR15,DBLP:conf/cvpr/HeZRS16,DBLP:conf/cvpr/HuangLMW17,chollet2017xception,DBLP:conf/icml/TanL19,DBLP:conf/iclr/DosovitskiyB0WZ21}. However, designing tailor-made highly performant CNNs for a given task is extremely difficult, as the required design choices intrinsic to the architectures, layer combination and training requires extensive architecture engineering. Thus, there is a growing interest in Neural Architecture Search (NAS) to automate architecture engineering and design.

NAS has successfully been successfully applied in the task of designing different types of neural network's architectures \citep{DBLP:journals/corr/abs-1905-01392}, specially for image and text problems \citep{elsken2019neural,DBLP:journals/corr/abs-1905-01392}. These methods are commonly composed of three components, being the first the search space, which specifies the possible operations to be sampled and their connections, ultimately defining the type of architectures that the search method can generate. The second component is the search method, which represents the approach used to explore the search space and generate architectures. The most common approaches are reinforcement learning, evolutionary strategies and gradient-based methods, which commonly work by updating a controller to sample more efficient architectures based on the performance of the generated models. Finally, the performance estimation strategy defines how the generated architectures are evaluated. Thus, the goal of a NAS method is to, based on the search method, efficiently search a large set of possible networks to find an optimal architecture for a given problem. Despite the excellent results obtain by prominent NAS methods, the computational cost of most approaches is high, which in some cases can be in the order of months of GPU computation \citep{DBLP:conf/iclr/ZophL17,liu2018progressive,Zoph_2018}. To mitigate this, interesting approaches focus on a cell-based design, where NAS methods design small cells that are replicated through an outer-skeleton, thus alleviating the complexity of the search space \citep{DBLP:conf/iclr/Shu0C20,DBLP:conf/iclr/ZophL17,Zoph_2018,carlucci2019manas}. More, several performance estimation strategies have been proposed to reduce the time constraint of NAS methods, by mainly conducting low-fidelity estimates, learning curve extrapolations, statistical approaches \citep{DBLP:journals/corr/abs-2104-01177,elsken2019neural} or by proposing one-shot methods, where the weights of the generated models are inherited \citep{DBLP:conf/iclr/LiuSY19,pham2018efficient,DBLP:conf/iclr/XuX0CQ0X20}. Searching through extensive search spaces is highly complex, even when there is some prior knowledge about the space. It has been shown that some of the most prominent NAS methods fail to generalise to new datasets due to converging extremely fast to local minima \citep{Dong2020NAS-Bench-201,Yang2020NAS}. The most reliable approach to obtain information about the search space while searching is to fully train generated architectures and optimise the search based on the most performant ones. However, this is costly, and results are highly dependant on the training schemes and initialisation setups. Therefore, zero-proxy estimators present an attractive solution, where statistics are drawn from the generated architectures to score them at initialisation stage, thus requiring no training \citep{epenas, mellor2020neural}. These methods are time efficient and capable of performing good correlations between the score and respective accuracies when the architectures are trained.

This paper proposes G-EA, an evolutionary NAS method that leverages zero-proxy estimation to guide the search. By using an evolutionary strategy, where operations can be mutated and younger architectures are prefered, G-EA forces an exploitation of the most performant networks, and an exploration of the search space by conducting mutations. More, we solve the problem of conducting full evaluation of the generated networks to obtain knowledge about the search space, by generating several architectures in each generation, where all are evaluated at initialisation stage using a zero-proxy estimator and only the highest scoring network is trained and kept for the next generation. By doing so, G-EA is capable of continuously extracting knowledge about the search space without compromising the search, resulting in state-of-the-art results in NAS-Bench-201 search space in CIFAR-10, CIFAR-100 and ImageNet16-120.

Our contributions can be summarized as:
\begin{itemize}
    \item We propose a novel guided NAS method based on evolutionary strategies and zero-proxy estimation to generate image classifier architectures - Convolutional Neural Networks.
    \item We empirically show that guided mechanisms can be used to improve the generated models performance without compromising time efficiency. Also, we detail the algorithm, emphasizing the accessible transferability of the guiding mechanism.
    \item We achieve state-of-the-art results in the NAS-Bench-201 search space, in all datasets: CIFAR-10, CIFAR-100 and ImageNet16-120.
\end{itemize}

\section{Related Work}
NAS was initially proposed as a Reinforcement Learning (RL) problem, where a controller is trained based on the generated architecture's performances to sample more efficient ones \citep{DBLP:conf/iclr/ZophL17}. Follow-up approaches focused on improving the overall performance, and the computation required to frame NAS as a RL problem by proposing the use of different learning strategies, distributed computing, and novel incremental sampling strategies \citep{zhong2018practical,elsken2019neural,DBLP:journals/corr/abs-1905-01392}. ENAS \citep{pham2018efficient}, showed that RL could be used to perform NAS in a reasonable time-frame by training a controller to discover architectures through optimal subgraph search within a large computational graph, requiring only a few computational days. DARTS, proposed the use of gradient-based approaches to generate architectures by performing a continuous relaxation of the parameters using a bi-level gradient optimization, resulting in the generation of competitive networks in a few GPU days \citep{DBLP:conf/iclr/LiuSY19}. These methods served as basis for follow-up weight-sharing NAS methods and one-shot models \citep{carlucci2019manas,dong2019searching,li2020random,DBLP:conf/iccv/Dong019a,zela2020understanding,DBLP:conf/iclr/XuX0CQ0X20}. Evolutionary strategies are also a common approach for NAS, which takes inspiration from biologic systems in order to generate architectures through a set of mutation operations. NEAT was the first evolutionary method to evolve simple neural networks \citep{stanley2002evolving}, which served as base and inspiration for methods that evolve deeper architectures where parent architectures have their parameters mutated to force evolution towards better performances \citep{DBLP:conf/icml/RealMSSSTLK17,DBLP:conf/iclr/ElskenMH19}. REA, is one of the most prominent approaches, in which the evolutionary strategy evolves architectures through operation and hidden states mutations, and also employs a tournament selection that favours younger architectures \citep{DBLP:conf/aaai/RealAHL19}.

Guiding mechanisms have been proposed to improve NAS. PNAS introduced a consortium learning to the search, where the design of architectures is gradual, based on the evaluation of increasingly larger networks \citep{liu2018progressive}. This approach allowed the method to be progressively guided through the search space, avoiding the need to train bad networks due to the estimation of the performance by a predictor network. However, this method still required immense computation. NPENAS guides an evolutionary search by proposing two predictors: a graph-based uncertainty estimation network and a performance predictor. NPENAS achieves a mean accuracy on NAS-Bench-201 CIFAR-10 of 91.07\% \citep{DBLP:journals/corr/abs-2003-12857}. In \citep{Bashivan_2019_ICCV}, the authors evaluate the similarity of the internal activations of generated architectures against a known one, e.g., ResNet, via representational similarity analysis to obtain knowledge regarding the search. \citep{Yu_2021_CVPR} proposes the use of landmark architecture's evaluation to regularize the ranking of child networks in super-net settings, thus guiding the search towards a better ranking correlation between stand-alone networks and the super-net ranking.

In this work, we propose a guided evolutionary method that is inspired by the findings that show that evolving architectures is an efficient approach for NAS, and that zero-proxy estimators provide a reasonably good and extremely fast scoring of untrained networks \citep{DBLP:conf/aaai/RealAHL19,epenas,DBLP:journals/corr/abs-2104-01177,DBLP:journals/corr/abs-1905-01392}. By coupling a zero-proxy estimator as a guiding mechanism to the search method, we force further exploitation of settings that are favourable to the architectures being generated, and, at the same time, also allows the exploration of the search space efficiently, by evaluating thousands of networks, providing information to guide the search.


\begin{algorithm}[!t]
\caption{Guided Evolution}
\label{guided_evol_alg}
\begin{algorithmic}
\State $population \gets $ empty queue  \Comment Population.
\State $history \gets \varnothing$  \Comment Models history.
\While{$\left\vert{population}\right\vert < C$}  \Comment Initialize population.
    \State $model.arch \gets \Call{RandomArchitecture}{ }$
    \State $model.accuracy \gets \Call{ZeroProxy}{model.arch}$
    \State add $model$ to right of $population$ \Comment Add model to the end, forcing age
\EndWhile
\State drop the $C-(C-P)$ worst individuals from $population$ 

\For{$model~\in~population$}
    \State $model.accuracy \gets \Call{TrainAndEval}{model.arch}$
    \State add $model$ to history
\EndFor

\While{$\left\vert{history}\right\vert < C $}  \Comment Evolve for $C$ cycles.
    \State $sample \gets \varnothing$  \Comment Parent candidates.
    \While{$\left\vert{sample}\right\vert < S$}
        \State $candidate \gets$ random model from $population$  \Comment Sampling with replacement.
        \State add $candidate$ to $sample$
    \EndWhile
    \State $parent \gets$ highest-accuracy model in $sample$
    
    \State $generation \gets $ empty list  \Comment Population.
    \While{$\left\vert{generation}\right\vert < P$}
        \State $child.arch \gets \Call{Mutate}{parent.arch}$
        \State $child.accuracy \gets \Call{ZeroProxy}{model.arch}$
        \State add $child$ to $generation$
    \EndWhile
    
    \State $top\_child \gets$ highest-performant model in $generation$
    \State $top\_child.accuracy \gets \Call{TrainAndEval}{model.arch}$
    \State add $top\_child$ to right of $population$
    \State add $top\_child$ to $history$
    \State remove $dead$ from left of $population$  \Comment Oldest model.
    \State discard $dead$
\EndWhile
\State \Return highest-accuracy model in $history$ \Comment Most performant model.
\end{algorithmic}
\end{algorithm}
\section{Proposed Method}
\label{sec:proposed}


The goal of NAS algorithms is to find an optimal architecture $a^*$ from the space of architectures $\mathcal{A}$, $a^*\in\mathcal{A}$, that maximizes an objective function $\mathcal{O}$. In this paper, we propose G-EA, which frames NAS as an optimization problem where an evolutionary strategy evolves architectures $a\in\mathcal{A}$ based on mutations and guided evolution. 

In the following sections, we detail G-EA and the zero-proxy estimator leveraged to create the guiding mechanism.

\subsection{Search Method}
G-EA is summarised in Algorithm \ref{guided_evol_alg}. In detail, G-EA starts by randomly generating $C$ architectures from the search space of possible architectures, $\mathcal{A}$. The architectures that belong to the search space have equal probabilities of being randomly sampled. Sampled architectures are then evaluated using a zero-proxy estimator that scores the architectures at initialisation stage, without requiring any training (the zero-proxy estimation mechanism is detailed in section \ref{subsec:zeroproxy-epenas}). Then, from the $C$ scored networks, only the top $P$ scoring architectures are added to the population and trained to extract their fitness, $f$, which is the validation accuracy. By scoring $C$ networks at initialisation stage, G-EA acquires knowledge regarding the search space, which is then exploited by selecting the top performant architectures, thus guiding the upcoming search by weeding out bad architectures.

\begin{wrapfigure}{h}{0.5\textwidth}
    \includegraphics[width=0.5\textwidth]{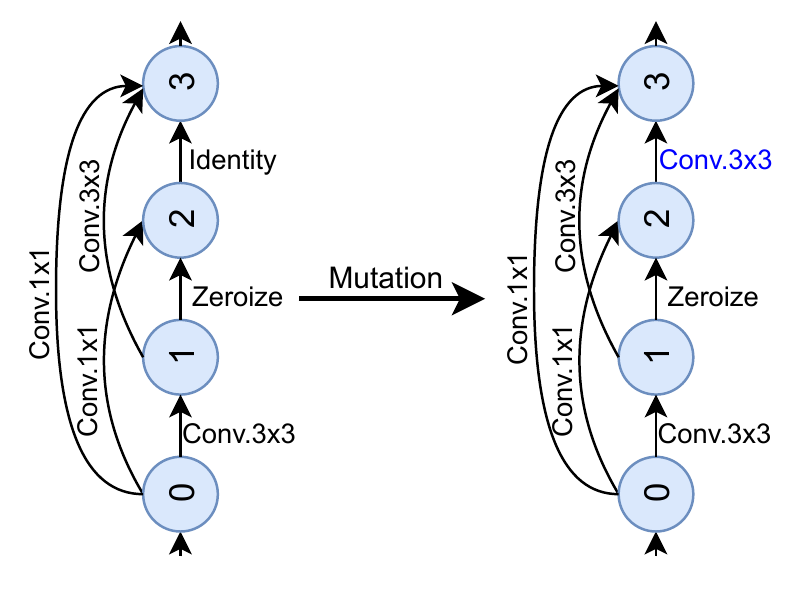}
    \caption{Representation of mutating an operation using NAS-Bench-201 search space. \label{fig:mutation}}
\end{wrapfigure}

Once the initial population is defined, the evolution takes place for $C$ cycles. At each iteration, the first step is to randomly and uniformly sample $S$ architectures from the population. Then, the architecture with the highest fitness score, $f$, is selected to be the parent of the next generation (cycle). To generate new architectures, G-EA performs a mutation over the parent architecture. The mutation works by randomly changing one operation of the architecture by another from the pool of operations. An example of a mutation using the NAS-Bench-201 search space is visually represented in Fig. \ref{fig:mutation}. $P$ new architectures are generated at each cycle by performing operation mutations over the selected parent, which are then scored using the zero-proxy estimator. The highest-scoring network is kept and added to the population after evaluating its fitness. By evaluating $P$ architectures, the search method can find which are the best directions to evolve the parent in the space. This allows the method to be guided through a complex space without jeopardizing the time required to perform the evolution or the search method's complexity. When the new architecture is added to the population, the oldest architecture is removed and discarded, thus forcing exploration of the search space by favouring younger architectures that represent new settings evolved by prior acquired knowledge.

Inherently, higher $P$ values represent a higher degree of exploration of the search space, while higher $S$ values represent higher exploitation by increasing the probability of the best architectures in the population being selected as parents for the next generation.

\subsection{Zero-proxy Estimator}
\label{subsec:zeroproxy-epenas}
To score networks at initialisation stage to aid in the guiding mechanism of the evolution, we use a zero-proxy estimator based on Jacobian covariance. This allows us to quickly evaluate if a network is good without requiring any training, thus allowing the selection of a generated network to be added to the population with more confidence that the search is being correctly guided to good spaces. To do this, we can define a linear mapping, $w_i=f(\mathbf{x}_i)$, which maps the input $\mathbf{x}_i  \in \mathbb{R}^{D}$, through the network, $f(\mathbf{x}_i)$, where $\mathbf{x}_i$ represents an image that belongs to a batch $\mathbf{X}$, and $D$ is the input dimension \citep{epenas}. Then, the Jacobian of the linear map can be computed using:
\begin{equation}
    \mathbf{J}_i = \frac{\partial f(\mathbf{x}_i)}{\partial \matr{x}_{i}}
\end{equation}


This allows us to evaluate a network behaviour for different images by calculating the Jacobian $\mathbf{w}_i$ for different data points, $f(\mathbf{x}_i)$, of a single batch $\mathbf{X}$, $i \in 1, \cdots, N$:
\begin{equation}
\matr{J} = 
\begin{pmatrix}
\frac{\partial f(\mathbf{x}_1)}{\partial \matr{x}_{1}} & \frac{\partial f(\mathbf{x}_2)}{\partial \matr{x}_{2}} & \cdots & \frac{\partial f(\mathbf{x}_N)}{\partial \matr{x}_{N}} \\
\end{pmatrix}^{\top}
\label{eq:J}
\end{equation}

$\matr{J}$ then contains information about the network output with respect to the input for several images. We can split this into classes and evaluate how an architecture models complex functions at the initialisation stage and its effect on images that belong to the same class. To do that, we split $\matr{J}$ into several sets, where each set, $\mathbf{M}_k$, contains all $\mathbf{J}_i$ that belong to the same class $k$. Then, we can calculate a per-class correlation matrix, $\matr{\Sigma}_{\mathbf{M}_k}$, using the obtained sets, $\mathbf{M}_k$, where $k=1,...K$.

Individual correlation matrices provide information about how a single architecture treats images for each class. However, different correlation matrices might yield different sizes, as the number of images per class differ. To be able to compare different correlation matrices, they are individually evaluated:
\begin{equation}
    \matr{E}_k =  
    \begin{cases}
        \sum_{i=1}^{N}\sum_{j=1}^{N} log(|({{\matr{\Sigma}_\mathbf{M}}_k})_{i,j}|+t), & \text{if }K \leq \tau\\\\
        \frac{\sum_{i=1}^{N}\sum_{j=1}^{N} log(|({{\matr{\Sigma}_\mathbf{M}}_c})_{i,j}|+t)}{||{{\matr{\Sigma}_\mathbf{M}}_k}||}, & \text{otherwise}
    \end{cases}
\end{equation}

\noindent where $t$ is a small-constant with the value of $1\times10^{-5}$, and $K$ is the number of classes in batch $\mathbf{X}$, and $||.||$ represents the size of the set X.

Finally, an architecture is scored based on the individual evaluations of the correlation matrices by:

\begin{equation}
    z =  
    \begin{cases}
        \sum_{w=1}^{K} |\matr{e}_w|, & \text{if }K \leq \tau\\\\
        \frac{\sum_{i=1}^{K}\sum_{j=i+1}^{K} |\matr{e}_i - \matr{e}_j|}  {||\matr{e}||} , & \text{otherwise}
    \end{cases}
\end{equation}

\noindent where $\matr{e}$ is a vector that contains all the correlation matrices' scores. The final score is dependant on the number of classes present in $\mathbf{X}$, as data sets with a higher number of classes commonly have more noise, which is mitigated by conducting a normalized pair-wise difference. In our experiments, we empirically defined $\tau=100$, based on the search space and data sets used.

We can then use $z$ to rank generated architectures, providing an efficient mechanism of differentiating between bad and good architectures.

\section{Experiments}
\label{sec:experiments}

\subsection{Search Space}
To evaluate the proposed method, we used the NAS-Bench-201 tabular benchmark \citep{Dong2020NAS-Bench-201}. NAS-Bench-201 fixes the search space as a cell-based design with 5 operations: zeroize, skip connection, $1\times1$ convolution, $3\times3$ convolution, and $3\times3$ average pooling layer. The cell design comprises six edges and four nodes, where an edge represents a possible operation through two nodes. By fixing the cell size and the operation pool, there are $5^6 = 15625$ possible cells in this search space. To form entire networks, the cells are replicated in an outer-defined skeleton. More, NAS-Bench-201 provides information regarding the training and performance of all possible networks in the search space in three datasets: CIFAR-10, CIFAR-100 and ImageNet16-120, thus allowing a quick prototyping and a controlled setting that allows different NAS methods to be fairly compared, as they are forced to use the search space, training procedures and hyper-parameters.

\subsection{Results and Discussion}

\begin{figure}[!t]
\begin{subfigure}[b]{0.326\textwidth}
  \centering
  \includegraphics[width=1\textwidth]{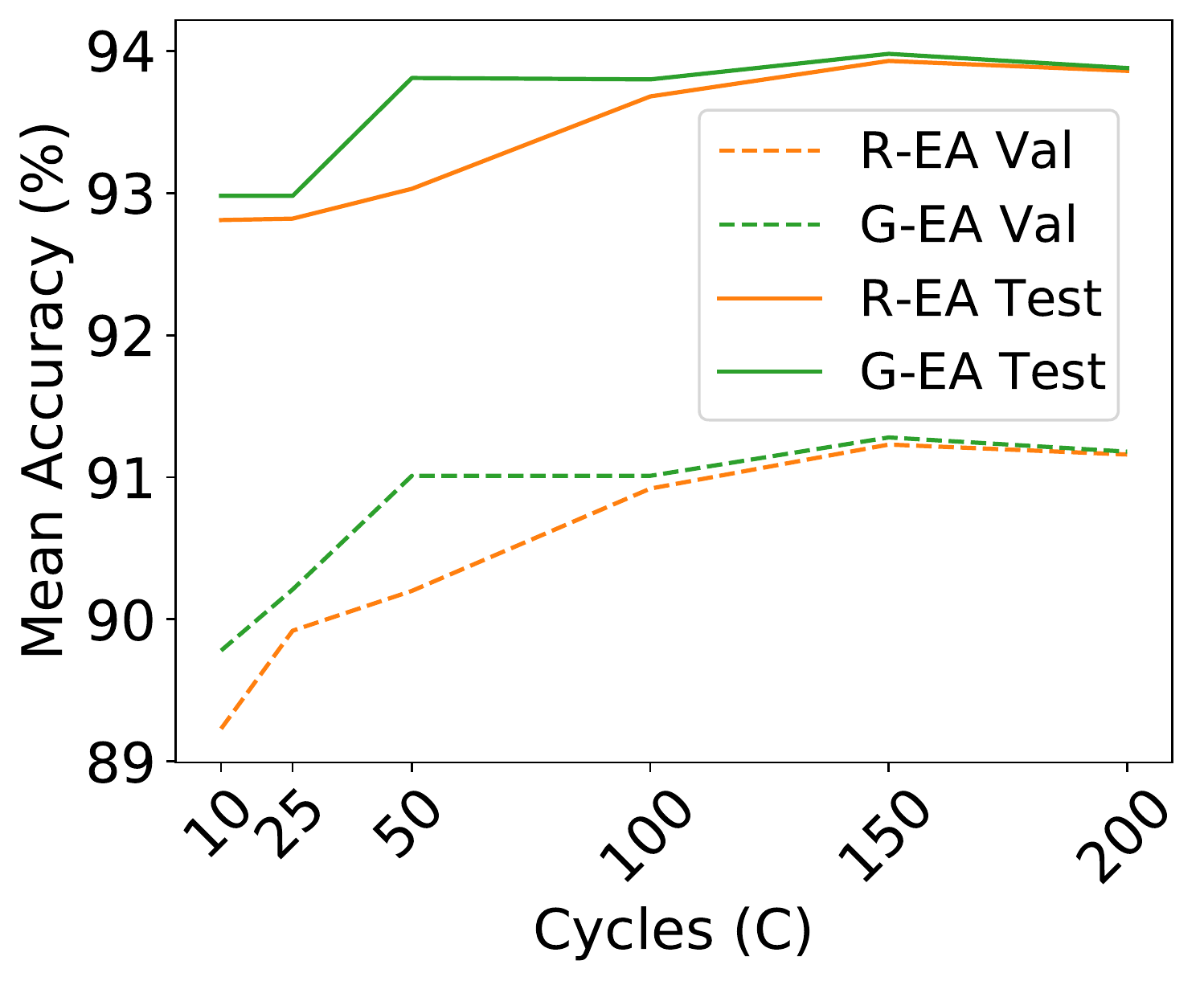}
  \caption{CIFAR-10}
  \label{fig:sfig2}
\end{subfigure}
\begin{subfigure}[b]{.31\textwidth}
  \centering
  \includegraphics[width=1\textwidth]{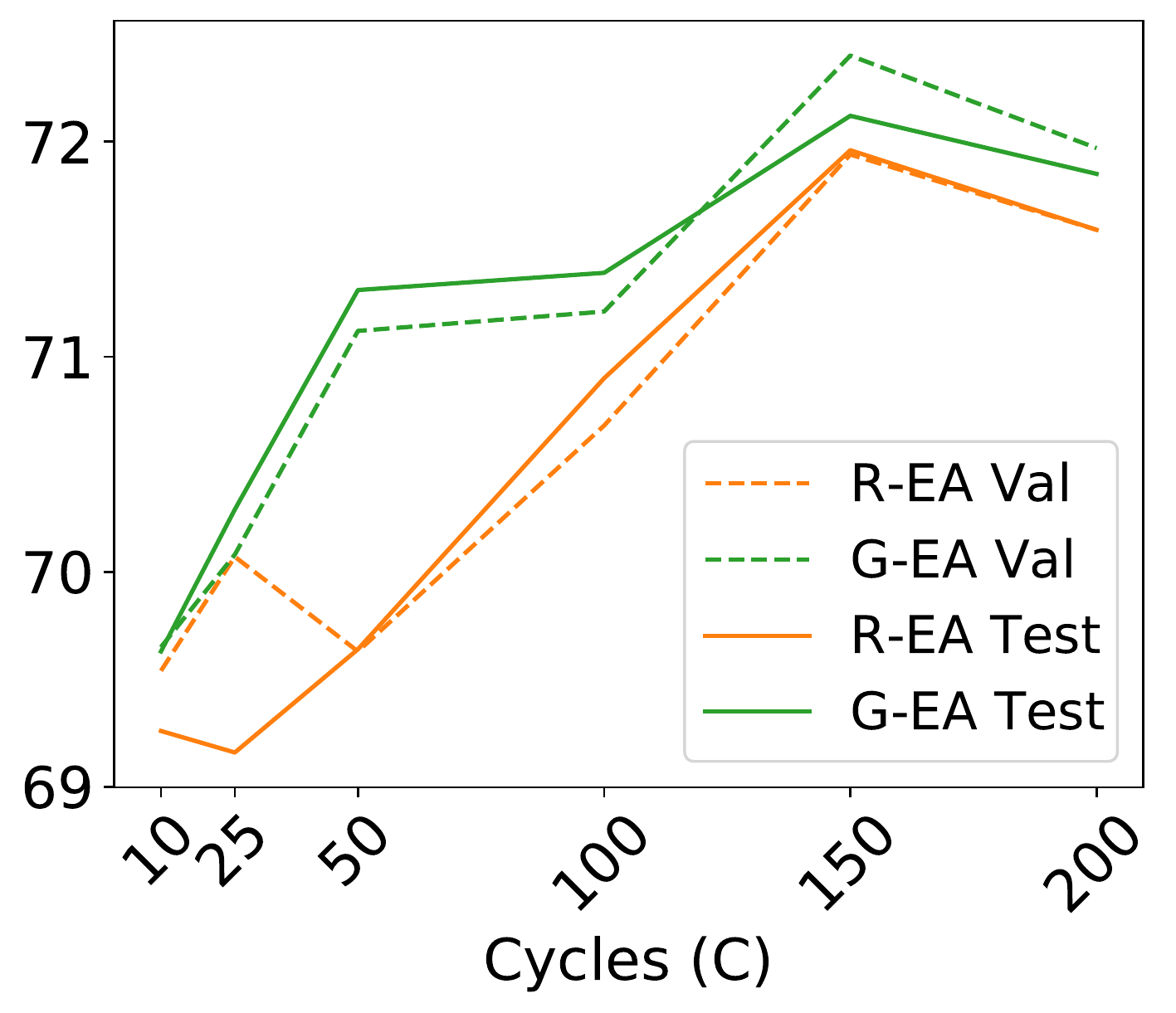}
  \caption{CIFAR-100}
  \label{fig:sfig3}
\end{subfigure}
\begin{subfigure}[b]{.31\textwidth}
  \centering
  \includegraphics[width=1\textwidth]{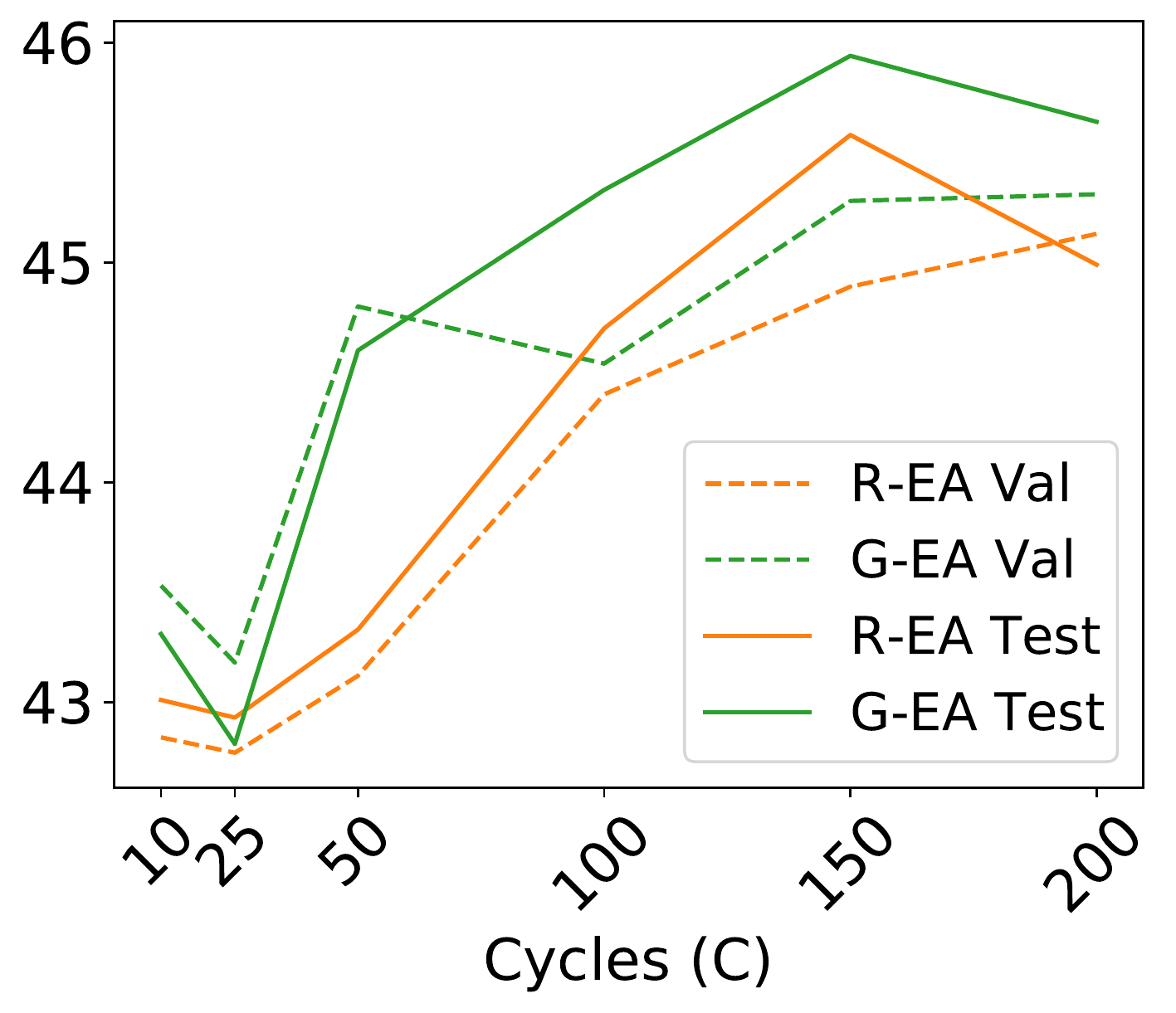}
  \caption{ImageNet16-120}
  \label{fig:sfig4}
\end{subfigure}
\caption{Mean accuracy over 10 runs of the proposed method, G-EA, and direct comparison with R-EA for different cycles ($C$) across CIFAR-10, CIFAR-100 and ImageNet16-120 data sets.}
\label{fig:cimportante}
\end{figure}

First, we evaluate the importance of the number of generations/cycles, $C$. This parameter inherently defines the time required for the search procedure. Higher $C$ values will take longer to finish. More, $C$ defines the number of architectures that are evaluated - $C^P$ architectures ($P$ per cycle) are generated and evaluated using the zero-proxy estimation method to provide information about the search space, from which, $C$ architectures ($1$ per cycle) are selected and trained. The results from this experiment can be seen in Fig \ref{fig:cimportante}, where a direct comparison with R-EA is also provided. The results are expressed as the mean accuracy over 10 runs, obtained by the best architecture found by each method both in the validation and test sets. In this experiment, the $P/S$ used to allow a fair comparison was set to $P/S=5/2$, following the common settings used and extensively evaluated by prior works \citep{Dong2020NAS-Bench-201,DBLP:conf/aaai/RealAHL19}. For our proposed method, G-EA, $P$ value means that at any given time of the search, the population is equal to $5$ architectures, and that from the pool of parents, $S$, $2$ architectures are sampled with replacement in order to elect the parent of the generated architectures at a given cycle. Denote that the sampled parent generates $P$ architectures through mutation per cycle, which are evaluated using the zero-proxy estimator, wherein the top scoring architecture is selected to integrate the population. By selecting $S>1$ architectures to have the opportunity of being a parent, we are leveraging the intrinsic exploitation characteristics of the evolutionary strategy, while by generating $P$ architectures, we are forcing an exploitation that guides the search more effectively.

\begin{table}[!t]
	\caption{Comparison of manually designed networks and several search methods evaluated using the NAS-Bench-201 benchmark. Performance is shown in terms of accuracy with mean$\pm$std, on CIFAR-10, CIFAR-100 and ImageNet-16-120. Search times are the mean time required to search for cells in CIFAR-10. Search time includes the time taken to train networks as part of the process where applicable. Table adapted from \citep{Dong2020NAS-Bench-201,epenas,mellor2020neural}.\label{table:benchmarkinggea}}
	\footnotesize
	\setlength{\tabcolsep}{3pt}
	\setlength{\arrayrulewidth}{2pt}
	
	\renewcommand{\arraystretch}{1.25}
	\resizebox{\textwidth}{!}{%
		\begin{tabular}{@{}lrllcllcll@{}} \hline  
			\multirow{2}{*}{Method} & \multirow{2}{*}{\shortstack{Search \\Time (s)}}  & \multicolumn{2}{c}{CIFAR-10} & \phantom{} & \multicolumn{2}{c}{CIFAR-100} & \phantom{} & \multicolumn{2}{c}{ImageNet-16-120} \\
			\cmidrule{3-4} \cmidrule{6-7} \cmidrule{9-10}
			& & \multicolumn{1}{c}{validation} & \multicolumn{1}{c}{test} && \multicolumn{1}{c}{validation} & \multicolumn{1}{c}{test} && \multicolumn{1}{c}{validation} & \multicolumn{1}{c}{test} \\
			\midrule

            \multicolumn{10}{c}{\textbf{Manually designed}}\\
			ResNet        & -  & \multicolumn{1}{c}{90.83} & \multicolumn{1}{c}{93.97} && \multicolumn{1}{c}{70.42} & \multicolumn{1}{c}{70.86} && \multicolumn{1}{c}{44.53} & \multicolumn{1}{c}{43.63} \\
            \midrule \midrule
			\multicolumn{10}{c}{\textbf{Weight sharing}}\\
			RSPS        & 7587  & 84.16$\pm$1.69 & 87.66$\pm$1.69 && 59.00$\pm$4.60 & 58.33$\pm$4.34 && 31.56$\pm$3.28 & 31.14$\pm$3.88 \\
			DARTS-V1    & 10890 & 39.77$\pm$0.00 & 54.30$\pm$0.00 && 15.03$\pm$0.00 & 15.61$\pm$0.00 && 16.43$\pm$0.00 & 16.32$\pm$0.00 \\
			DARTS-V2    & 29902 & 39.77$\pm$0.00 & 54.30$\pm$0.00 && 15.03$\pm$0.00 & 15.61$\pm$0.00 && 16.43$\pm$0.00 & 16.32$\pm$0.00 \\
			GDAS        & 28926 & 90.00$\pm$0.21 & 93.51$\pm$0.13 && 71.14$\pm$0.27 & 70.61$\pm$0.26 && 41.70$\pm$1.26 & 41.84$\pm$0.90 \\
			SETN        & 31010 & 82.25$\pm$5.17 & 86.19$\pm$4.63 && 56.86$\pm$7.59 & 56.87$\pm$7.77 && 32.54$\pm$3.63 & 31.90$\pm$4.07 \\
			ENAS        & 13315 & 39.77$\pm$0.00 & 54.30$\pm$0.00 && 15.03$\pm$0.00 & 15.61$\pm$0.00 && 16.43$\pm$0.00 & 16.32$\pm$0.00 \\

			\midrule \midrule
			\multicolumn{10}{c}{\textbf{Non-weight sharing}}\\
			RS        &  12000 & 90.93$\pm$0.36 & 93.70$\pm$0.36 && 70.93$\pm$1.09 & 71.04$\pm$1.07 && 44.45$\pm$1.10 & 44.57$\pm$1.25 \\
			REINFORCE &  12000 & 91.09$\pm$0.37 & 93.85$\pm$0.37 && 71.61$\pm$1.12 & 71.71$\pm$1.09 && 45.05$\pm$1.02 & 45.24$\pm$1.18 \\
			BOHB      &  12000 & 90.82$\pm$0.53 & 93.61$\pm$0.52 && 70.74$\pm$1.29 & 70.85$\pm$1.28 && 44.26$\pm$1.36 & 44.42$\pm$1.49 \\
			REA$\dagger$                  & 18577 &91.23$\pm$0.29  &93.93$\pm$0.31  &&71.94$\pm$1.24  & 71.96$\pm$1.19 &&44.89$\pm$1.00 & 45.58$\pm$1.01
			\\
			\textbf{G-EA (ours)}$\dagger$ & 18567 &\textbf{91.28}$\pm$\textbf{0.12}  &\textbf{93.98}$\pm$\textbf{0.18}  &&\textbf{72.40}$\pm$\textbf{0.41}  & \textbf{72.12}$\pm$\textbf{0.35} && \textbf{45.28}$\pm$\textbf{0.68}& \textbf{45.94}$\pm$\textbf{0.71}
			\\
			\bottomrule
		\end{tabular}
	}
	\scriptsize $\dagger$~Results of 10 runs using the same settings: $P/S/C=5/2/150$, using a single 1080Ti GPU.\\
\end{table}

From Fig. \ref{fig:cimportante}, it is possible to see that across all datasets, G-EA consistently outperforms R-EA, and is capable of converging to better results even with a low $C$. These results demonstrate that by providing a guided mechanism, the search method convergences more quickly to regions of the search space that contain better architectures, peaking at $C=150$ in these settings. Based on these results, in Table \ref{table:benchmarkinggea} we further compare G-EA using $P/S/C=5/2/150$ against other state-of-the-art methods on the NAS-Bench-201 search space, using as evaluation metrics the mean accuracy, standard deviation and search time, in seconds, across the 3 data sets. G-EA consistently outperforms both weight sharing and non-weight sharing NAS methods, achieving state-of-the-art results in all three data sets. Moreover, G-EA is extremely efficient in terms of search time, requiring only 0.2 GPU days to complete the search. Even though G-EA evaluates $C^P$ architectures with the zero-proxy estimator and further evaluates $C$ architectures by training them, it requires a similar search time as REA under the same settings, and considerably less than most weight sharing methods. Lower standard deviation also indicates that G-EA is precise and capable of generating high performant architectures. This is specially valid in ImageNet16-120, a data set with low resolution images and high levels of noise, in which G-EA considerably outperforms existing NAS methods. 

The obtained results show that evolutionary strategies coupled with a mechanism to quickly evaluate architectures to guide the search can achieve state-of-the-art results while still having competitive search times. Despite the complexity of search spaces and severe difficulty in obtaining their global information, the achieved results shed insights that guiding mechanisms powered by scoring architectures at initialisation stages give us the advantage of acquiring preliminary information regarding which direction should the search evolve to. Therefore, G-EA is capable of avoiding local minima and quickly converge to better results while still being capable of improving the time required by the search method.

\section{Conclusions}
This paper proposes G-EA, a guided evolution strategy for neural architecture search by leveraging zero-proxy estimation of untrained architectures. G-EA forces exploitation of the most performant networks by descendant generation and an exploration of the search space by conducting mutations. G-EA guides the evolution by exploring the search space by generating several architectures in each generation and having them evaluated at initialisation stage using a zero-proxy estimator, where only the highest-scoring network is trained and kept for the next generation. By generating several architectures from an existing architecture from the population at each generation, G-EA is capable of continuously extracting knowledge about the search space without compromising the search, resulting in state-of-the-art results in NAS-Bench-201 search space in CIFAR-10, CIFAR-100 and ImageNet16-120, with mean accuracies of 93.98\%, 72.12\% and 45.94\% respectively.

The simplicity of our approach allows it to easily be extended, where the search method is further improved by incorporating new regularisation and mutation mechanisms. Also, the components that compose the guiding mechanism can easily be transferred to other evolutionary algorithms, allowing existing NAS evolutionary methods to be further improved.

\begin{ack}
This work was supported by `FCT - Fundação para a Ciência e Tecnologia' through the research grants `2020.04588.BD' and `UI/BD/150765/2020', and partially supported by NOVA LINCS (UIDB/04516/2020) with the financial support of FCT, through national funds.
\end{ack}

{
\small
\bibliographystyle{abbrvnat}
\bibliography{mybibfile}

\begin{thebibliography}{39}
\providecommand{\natexlab}[1]{#1}
\providecommand{\url}[1]{\texttt{#1}}
\expandafter\ifx\csname urlstyle\endcsname\relax
  \providecommand{\doi}[1]{doi: #1}\else
  \providecommand{\doi}{doi: \begingroup \urlstyle{rm}\Url}\fi

\bibitem[Bashivan et~al.(2019)Bashivan, Tensen, and
  DiCarlo]{Bashivan_2019_ICCV}
P.~Bashivan, M.~Tensen, and J.~J. DiCarlo.
\newblock Teacher guided architecture search.
\newblock In \emph{Proceedings of the IEEE/CVF International Conference on
  Computer Vision (ICCV)}, October 2019.

\bibitem[Carlucci et~al.(2019)Carlucci, Esperan{\c{c}}a, Singh, Gabillon, Yang,
  Xu, Chen, and Wang]{carlucci2019manas}
F.~M. Carlucci, P.~M. Esperan{\c{c}}a, M.~Singh, V.~Gabillon, A.~Yang, H.~Xu,
  Z.~Chen, and J.~Wang.
\newblock Manas: Multi-agent neural architecture search.
\newblock \emph{arXiv preprint arXiv:1909.01051}, 2019.

\bibitem[Chollet(2017)]{chollet2017xception}
F.~Chollet.
\newblock Xception: Deep learning with depthwise separable convolutions.
\newblock In \emph{Proceedings of the IEEE conference on computer vision and
  pattern recognition}, pages 1251--1258, 2017.

\bibitem[Conneau et~al.(2017)Conneau, Schwenk, Barrault, and
  LeCun]{DBLP:conf/eacl/SchwenkBCL17}
A.~Conneau, H.~Schwenk, L.~Barrault, and Y.~LeCun.
\newblock Very deep convolutional networks for text classification.
\newblock In M.~Lapata, P.~Blunsom, and A.~Koller, editors, \emph{15th
  Conference of the European Chapter of the Association for Computational
  Linguistics, {EACL}}. Association for Computational Linguistics, 2017.

\bibitem[Deng et~al.(2014)Deng, Yu, et~al.]{deng2014deep}
L.~Deng, D.~Yu, et~al.
\newblock Deep learning: methods and applications.
\newblock \emph{Foundations and Trends in Signal Processing}, 2014.

\bibitem[Dong and Yang(2019{\natexlab{a}})]{DBLP:conf/iccv/Dong019a}
X.~Dong and Y.~Yang.
\newblock {One-Shot Neural Architecture Search via Self-Evaluated Template
  Network}.
\newblock In \emph{{ICCV}}. {IEEE}, 2019{\natexlab{a}}.

\bibitem[Dong and Yang(2019{\natexlab{b}})]{dong2019searching}
X.~Dong and Y.~Yang.
\newblock Searching for a robust neural architecture in four gpu hours.
\newblock In \emph{CVPR}, pages 1761--1770, 2019{\natexlab{b}}.

\bibitem[Dong and Yang(2020)]{Dong2020NAS-Bench-201}
X.~Dong and Y.~Yang.
\newblock {NAS-Bench-201: Extending the Scope of Reproducible Neural
  Architecture Search}.
\newblock In \emph{ICLR}, 2020.

\bibitem[Dosovitskiy et~al.(2021)Dosovitskiy, Beyer, Kolesnikov, Weissenborn,
  Zhai, Unterthiner, Dehghani, Minderer, Heigold, Gelly, Uszkoreit, and
  Houlsby]{DBLP:conf/iclr/DosovitskiyB0WZ21}
A.~Dosovitskiy, L.~Beyer, A.~Kolesnikov, D.~Weissenborn, X.~Zhai,
  T.~Unterthiner, M.~Dehghani, M.~Minderer, G.~Heigold, S.~Gelly, J.~Uszkoreit,
  and N.~Houlsby.
\newblock An image is worth 16x16 words: Transformers for image recognition at
  scale.
\newblock In \emph{9th International Conference on Learning Representations,
  {ICLR} 2021, Virtual Event, Austria, May 3-7, 2021}, 2021.

\bibitem[Elsken et~al.(2019{\natexlab{a}})Elsken, Metzen, and
  Hutter]{DBLP:conf/iclr/ElskenMH19}
T.~Elsken, J.~H. Metzen, and F.~Hutter.
\newblock Efficient multi-objective neural architecture search via lamarckian
  evolution.
\newblock In \emph{{ICLR}}, 2019{\natexlab{a}}.

\bibitem[Elsken et~al.(2019{\natexlab{b}})Elsken, Metzen, and
  Hutter]{elsken2019neural}
T.~Elsken, J.~H. Metzen, and F.~Hutter.
\newblock Neural architecture search: A survey.
\newblock \emph{Journal of Machine Learning Research}, 2019{\natexlab{b}}.

\bibitem[Garcia-Garcia et~al.(2018)Garcia-Garcia, Orts-Escolano, Oprea,
  Villena-Martinez, Martinez-Gonzalez, and Garcia-Rodriguez]{garcia2018survey}
A.~Garcia-Garcia, S.~Orts-Escolano, S.~Oprea, V.~Villena-Martinez,
  P.~Martinez-Gonzalez, and J.~Garcia-Rodriguez.
\newblock A survey on deep learning techniques for image and video semantic
  segmentation.
\newblock \emph{Applied Soft Computing}, 70:\penalty0 41--65, 2018.

\bibitem[Goodfellow et~al.(2016)Goodfellow, Bengio, Courville, and
  Bengio]{goodfellow2016deep}
I.~Goodfellow, Y.~Bengio, A.~Courville, and Y.~Bengio.
\newblock \emph{Deep learning}, volume~1.
\newblock MIT press Cambridge, 2016.

\bibitem[He et~al.(2016)He, Zhang, Ren, and Sun]{DBLP:conf/cvpr/HeZRS16}
K.~He, X.~Zhang, S.~Ren, and J.~Sun.
\newblock Deep residual learning for image recognition.
\newblock In \emph{{CVPR}}, 2016.

\bibitem[Huang et~al.(2017)Huang, Liu, van~der Maaten, and
  Weinberger]{DBLP:conf/cvpr/HuangLMW17}
G.~Huang, Z.~Liu, L.~van~der Maaten, and K.~Q. Weinberger.
\newblock Densely connected convolutional networks.
\newblock In \emph{{CVPR}}, 2017.

\bibitem[Khan et~al.(2020)Khan, Sohail, Zahoora, and Qureshi]{khan2020survey}
A.~Khan, A.~Sohail, U.~Zahoora, and A.~S. Qureshi.
\newblock A survey of the recent architectures of deep convolutional neural
  networks.
\newblock \emph{Artificial Intelligence Review}, 2020.

\bibitem[Krizhevsky et~al.(2012)Krizhevsky, Sutskever, and
  Hinton]{DBLP:conf/nips/KrizhevskySH12}
A.~Krizhevsky, I.~Sutskever, and G.~E. Hinton.
\newblock Imagenet classification with deep convolutional neural networks.
\newblock In \emph{Advances in Neural Information Processing Systems}, 2012.

\bibitem[Li and Talwalkar(2020)]{li2020random}
L.~Li and A.~Talwalkar.
\newblock Random search and reproducibility for neural architecture search.
\newblock In \emph{UAI}, pages 367--377. PMLR, 2020.

\bibitem[Liu et~al.(2018)Liu, Zoph, Neumann, Shlens, Hua, Li, Fei-Fei, Yuille,
  Huang, and Murphy]{liu2018progressive}
C.~Liu, B.~Zoph, M.~Neumann, J.~Shlens, W.~Hua, L.-J. Li, L.~Fei-Fei,
  A.~Yuille, J.~Huang, and K.~Murphy.
\newblock Progressive neural architecture search.
\newblock In \emph{Proceedings of the European conference on computer vision
  (ECCV)}, 2018.

\bibitem[Liu et~al.(2019)Liu, Simonyan, and Yang]{DBLP:conf/iclr/LiuSY19}
H.~Liu, K.~Simonyan, and Y.~Yang.
\newblock {{DARTS:} Differentiable Architecture Search}.
\newblock In \emph{{ICLR}}, 2019.

\bibitem[Lopes et~al.(2021)Lopes, Alirezazadeh, and Alexandre]{epenas}
V.~Lopes, S.~Alirezazadeh, and L.~A. Alexandre.
\newblock {EPE-NAS: Efficient Performance Estimation Without Training for
  Neural Architecture Search}.
\newblock In \emph{International Conference on Artificial Neural Networks
  (ICANN)}, 2021.

\bibitem[Mellor et~al.(2021)Mellor, Turner, Storkey, and
  Crowley]{mellor2020neural}
J.~Mellor, J.~Turner, A.~J. Storkey, and E.~J. Crowley.
\newblock {Neural Architecture Search without Training}.
\newblock In \emph{ICML}, 2021.

\bibitem[Pham et~al.(2018)Pham, Guan, Zoph, Le, and Dean]{pham2018efficient}
H.~Pham, M.~Guan, B.~Zoph, Q.~Le, and J.~Dean.
\newblock Efficient neural architecture search via parameters sharing.
\newblock In \emph{ICML}, 2018.

\bibitem[Real et~al.(2017)Real, Moore, Selle, Saxena, Suematsu, Tan, Le, and
  Kurakin]{DBLP:conf/icml/RealMSSSTLK17}
E.~Real, S.~Moore, A.~Selle, S.~Saxena, Y.~L. Suematsu, J.~Tan, Q.~V. Le, and
  A.~Kurakin.
\newblock Large-scale evolution of image classifiers.
\newblock In D.~Precup and Y.~W. Teh, editors, \emph{Proceedings of the 34th
  International Conference on Machine Learning, {ICML} 2017, Sydney, NSW,
  Australia, 6-11 August 2017}, volume~70 of \emph{Proceedings of Machine
  Learning Research}, pages 2902--2911. {PMLR}, 2017.

\bibitem[Real et~al.(2019)Real, Aggarwal, Huang, and
  Le]{DBLP:conf/aaai/RealAHL19}
E.~Real, A.~Aggarwal, Y.~Huang, and Q.~V. Le.
\newblock {Regularized Evolution for Image Classifier Architecture Search}.
\newblock In \emph{{AAAI}}, pages 4780--4789. {AAAI} Press, 2019.

\bibitem[Shu et~al.(2020)Shu, Wang, and Cai]{DBLP:conf/iclr/Shu0C20}
Y.~Shu, W.~Wang, and S.~Cai.
\newblock Understanding architectures learnt by cell-based neural architecture
  search.
\newblock In \emph{8th International Conference on Learning Representations,
  {ICLR} 2020, Addis Ababa, Ethiopia, April 26-30, 2020}. OpenReview.net, 2020.

\bibitem[Stanley and Miikkulainen(2002)]{stanley2002evolving}
K.~O. Stanley and R.~Miikkulainen.
\newblock Evolving neural networks through augmenting topologies.
\newblock \emph{Evolutionary computation}, 10\penalty0 (2):\penalty0 99--127,
  2002.

\bibitem[Szegedy et~al.(2015)Szegedy, Liu, Jia, Sermanet, Reed, Anguelov,
  Erhan, Vanhoucke, and Rabinovich]{DBLP:conf/cvpr/SzegedyLJSRAEVR15}
C.~Szegedy, W.~Liu, Y.~Jia, P.~Sermanet, S.~E. Reed, D.~Anguelov, D.~Erhan,
  V.~Vanhoucke, and A.~Rabinovich.
\newblock Going deeper with convolutions.
\newblock In \emph{{CVPR}}, 2015.

\bibitem[Tan and Le(2019)]{DBLP:conf/icml/TanL19}
M.~Tan and Q.~V. Le.
\newblock Efficientnet: Rethinking model scaling for convolutional neural
  networks.
\newblock In K.~Chaudhuri and R.~Salakhutdinov, editors, \emph{{ICML}}, 2019.

\bibitem[Wei et~al.(2020)Wei, Niu, Tang, and
  Liang]{DBLP:journals/corr/abs-2003-12857}
C.~Wei, C.~Niu, Y.~Tang, and J.~Liang.
\newblock {NPENAS:} neural predictor guided evolution for neural architecture
  search.
\newblock \emph{CoRR}, abs/2003.12857, 2020.

\bibitem[White et~al.(2021)White, Zela, Ru, Liu, and
  Hutter]{DBLP:journals/corr/abs-2104-01177}
C.~White, A.~Zela, B.~Ru, Y.~Liu, and F.~Hutter.
\newblock How powerful are performance predictors in neural architecture
  search?
\newblock \emph{CoRR}, abs/2104.01177, 2021.

\bibitem[Wistuba et~al.(2019)Wistuba, Rawat, and
  Pedapati]{DBLP:journals/corr/abs-1905-01392}
M.~Wistuba, A.~Rawat, and T.~Pedapati.
\newblock A survey on neural architecture search.
\newblock \emph{CoRR}, abs/1905.01392, 2019.

\bibitem[Xu et~al.(2020)Xu, Xie, Zhang, Chen, Qi, Tian, and
  Xiong]{DBLP:conf/iclr/XuX0CQ0X20}
Y.~Xu, L.~Xie, X.~Zhang, X.~Chen, G.~Qi, Q.~Tian, and H.~Xiong.
\newblock {PC-DARTS:} partial channel connections for memory-efficient
  architecture search.
\newblock In \emph{8th International Conference on Learning Representations,
  {ICLR} 2020, Addis Ababa, Ethiopia, April 26-30, 2020}. OpenReview.net, 2020.

\bibitem[Yang et~al.(2020)Yang, Esperança, and Carlucci]{Yang2020NAS}
A.~Yang, P.~M. Esperança, and F.~M. Carlucci.
\newblock Nas evaluation is frustratingly hard.
\newblock In \emph{ICLR}, 2020.

\bibitem[Yu et~al.(2021)Yu, Ranftl, and Salzmann]{Yu_2021_CVPR}
K.~Yu, R.~Ranftl, and M.~Salzmann.
\newblock Landmark regularization: Ranking guided super-net training in neural
  architecture search.
\newblock In \emph{Proceedings of the IEEE/CVF Conference on Computer Vision
  and Pattern Recognition (CVPR)}, pages 13723--13732, June 2021.

\bibitem[Zela et~al.(2020)Zela, Elsken, Saikia, Marrakchi, Brox, and
  Hutter]{zela2020understanding}
A.~Zela, T.~Elsken, T.~Saikia, Y.~Marrakchi, T.~Brox, and F.~Hutter.
\newblock Understanding and robustifying differentiable architecture search.
\newblock In \emph{ICLR}, 2020.

\bibitem[Zhong et~al.(2018)Zhong, Yan, Wu, Shao, and Liu]{zhong2018practical}
Z.~Zhong, J.~Yan, W.~Wu, J.~Shao, and C.-L. Liu.
\newblock Practical block-wise neural network architecture generation.
\newblock In \emph{CVPR}, 2018.

\bibitem[Zoph and Le(2017)]{DBLP:conf/iclr/ZophL17}
B.~Zoph and Q.~V. Le.
\newblock Neural architecture search with reinforcement learning.
\newblock In \emph{{ICLR}}, 2017.

\bibitem[Zoph et~al.(2018)Zoph, Vasudevan, Shlens, and Le]{Zoph_2018}
B.~Zoph, V.~Vasudevan, J.~Shlens, and Q.~V. Le.
\newblock Learning transferable architectures for scalable image recognition.
\newblock \emph{CVPR}, Jun 2018.

\end{thebibliography}

}




\end{document}